\pdfoutput=1

\documentclass[11pt]{article}

\usepackage[]{emnlp2021}

\usepackage{times}
\usepackage{latexsym}
\usepackage{graphicx}
\usepackage{hyperref}
\usepackage{tabularx}
\usepackage{todonotes}

\usepackage[T1]{fontenc}

\usepackage[utf8]{inputenc}

\usepackage{microtype}

\title{Is Stance Detection Topic-Independent and Cross-topic Generalizable? \\ -- A Reproduction Study
}
\author{
\begin{tabular}{@{}c@{}}
Myrthe Reuver$^1$ \qquad 
  Suzan Verberne$^3$ \qquad 
  Roser Morante$^{1}$ \qquad 
  Antske Fokkens$^1$$^,$$^2$ \\
\end{tabular}\\

$^1$Computation Linguistics and Text Mining Lab, Vrije Universiteit Amsterdam \\ $^2$ Dept. of Mathematics and Computer Science, Eindhoven University of Technology \\ $^3$ Leiden Institute of Advanced Computer Science, Leiden University\\
\texttt{\{myrthe.reuver,antske.fokkens,r.morantevallejo\}@vu.nl}\\ \texttt{s.verberne@liacs.leidenuniv.nl}}

\begin{document}
\maketitle

\begin{abstract}
Cross-topic stance detection is the task to automatically detect stances (pro, against, or neutral) on unseen topics. We successfully reproduce state-of-the-art cross-topic stance detection work \cite{reimers_classification_2019}, and systematically analyze its reproducibility. Our attention then turns to the cross-topic aspect of this work, and the specificity of topics in terms of vocabulary and socio-cultural context. 
We ask: To what extent is stance detection topic-independent and generalizable across topics? We compare the model's performance on various unseen topics, and find topic (e.g.\ abortion, cloning), class (e.g.\ pro, con), and their interaction affecting the model's performance. We conclude that investigating performance on different topics, and addressing topic-specific vocabulary and context, is a future avenue for cross-topic stance detection.
\end{abstract}

\section{Introduction}\label{sec:introduction}
(Online) debate has long been studied and modelled by computational linguistics with argument mining tasks such as stance detection. Stance detection is the task of automatically identifying the stance (agreeing, disagreeing, and/or neutral) of a text towards a debated topic or issue \cite{StanceSurvey, schiller_stance_2021}.\footnote{There is a wide array of datasets, definitions, and operationalizations of stance detection and classification, and recently \citet{schiller_stance_2021} gave a great overview in their Section 2, as do \citet{StanceSurvey} in their survey.} Its use-cases increasingly relate to online information environments and societal challenges, such as argument search \cite{stab2018argumentext}, fake news identification \cite{hanselowski2018retrospective}, or diversifying stances in a news recommender \cite{reuver-etal-2021-nlp}. 

Cross-topic stance detection models should thus be able to deal with the quickly changing landscape of (online) public debate, where new topics and issues appear all the time. As \citet{schlangen-20191-natural} described in his recent paper on natural language processing (NLP) methodology, generalization is a main goal of computational linguistics. A computational model (e.g.\ a stance detection model) should learn task capabilities beyond one set of datapoints, in our case: beyond one debate topic. 

Cross-topic stance detection is especially challenging because generalization to a new discussion topic is not trivial. Expressing stances is inherently socio-cultural behavior \cite{du2007stance}, where social actors place themselves and targets on dimensions in the socio-cultural field. This also comes with very topic-specific word use \cite{somasundaran-wiebe-2009-recognizing, wei2019modeling}. For instance, an \textit{against} abortion argument might be expressed indirectly 
with a `pro-life' expression, and someone aware of the socio-cultural context of this debate will be able to recognize this. Knowledge from other debate topics such as \textit{gun control} may not be useful, since the debate strategies might change per topic. 
Despite these fundamental challenges, pre-trained Transformer models show promising results on cross-topic argument classification \cite{reimers_classification_2019, schiller_stance_2021}.

In this paper, we investigate the ability of cross-topic stance detection approaches to generalize to different debate topics. Our question is: \textit{To what extent is stance detection topic-independent and generalizable across topics?}

Our contributions are threefold. We first complete a reproduction of state-of-the-art cross-topic stance detection work \cite{reimers_classification_2019}, as reproduction has repeatedly shown to be relevant for NLP \cite{fokkens2013offspring, cohen2018three,  belz-etal-2021-systematic}.  The reproduction is largely successful: we obtain similar numeric results. 
Secondly, we investigate the topic-specific performance of this model, and conclude that BERT's performance fluctuates on different topics. Additionally, we find that a bag-of-words-based SVM model can rival its performance for some topics. Thirdly, we relate this to the nature of the stance detection modelling task, which is inherently more connected to socio-cultural aspects and topic-specific differences than related tasks such as sentiment analysis. 

This paper is organized as follows. Section~\ref{sec:Background} discusses earlier work on stance detection, and specifically generalizability across topics. Section~\ref{sec:Repro} presents the reproduction results. Section~\ref{sec:Topics} adds additional, topic-specific analyses of the classification performance and a bag-of-words-based model to find topic-(in)dependent features. This is followed by our conclusions in Section~\ref{sec:Discussion}.

\section{Background}\label{sec:Background}
\subsection{Definition of Stance Detection}
Stance detection is a long-established task in computational linguistics. \citet{StanceSurvey} identify its most commonly used task definition: “For an input in the form of a piece of text and a target pair, stance detection is a classification problem where the stance of the author of the text is sought in the form of a category label from this set: {Favor, Against, Neither}.” \cite[p. 2]{StanceSurvey}.\footnote{We would like to note that the stance expressed in a text unit does not have to be the stance of an author, e.g.\ in cases where someone is writing a piece in which they express or quote someone else's opinion.} The number of stance classes can vary from 2 to 4, e.g.\ by adding `comment' and `query' next to `for' and `against' \cite{schiller_stance_2021}. \citet{StanceSurvey} emphasize that this computational definition is built upon the linguistic phenomenon of actors communicating their evaluation of targets, by which they place themselves and their targets on ``dimensions in the sociocultural field'' \cite[p. 163]{du2007stance}. Current work focuses mostly on debates deemed controversial in the U.S. socio-political domain, such as abortion and gun control.

\subsection{Prior work}
Early work on stance detection focused on parliamentary debates and longer texts \cite{thomas2006get}. 
Since \citet{mohammad-etal-2016-semeval}'s stance detection shared task, 
Twitter has attracted a lot of attention in NLP work on stance detection \cite{zhu2019hierarchical, darwish2020unsupervised, hossain2020covidlies}. Others addressed stance detection in the news domain, with (fake) news headlines \cite{ferreira2016emergent, hanselowski2018retrospective}, disinformation \cite{hardalov2021survey} and user comments on news websites \cite{bovsnjak2019data}.

Feature-based approaches have largely been replaced by end-to-end neural models. Stance detection has seen a performance increase due to pre-trained Transformer models such as BERT \cite{devlin-etal-2019-bert}. \citet{reimers_classification_2019} reported .20 point F1 improvement over an LSTM baseline with a pre-trained BERT model. Combining multiple stance detection datasets in fine-tuning such a pre-trained Transformer again led to a performance increase, though this model lacks robustness against slight test set manipulations \cite{schiller_stance_2021}.

\subsection{Generalization to new topics}
Recent work has specifically worked on identifying stances on topics not seen in training.
\citet{reimers_classification_2019} train their model on detecting stances and arguments for unseen topics. In their approach however, they treat all topics and stances on these topics as similar and comparable, and report one averaged evaluation metric over topics.

Earlier work \cite{somasundaran-wiebe-2009-recognizing} already established that 
ideological stances on topics deemed controversial, such as gay rights, are expressed in a topic-specific manner. Topic-specific features 
were more informative for SVM models than more topic-independent features.
 
In more recent work, \citet{wei2019modeling} instead specifically focus on how generalizable certain topics are for transferring knowledge to new topics on stance detection. Some Twitter discussion topics seem to share a latent, underlying topic (e.g.\ both feminism and abortion have the latent topic of equality). In a (latent) topic-enhanced multi-layer perceptron (MLP) model with RNN representation of the tweet, the model indeed uses shared vocabulary between the related topics. 

\citet{allaway-etal-2021-adversarial} notice that earlier work, when considering training on some topics and testing on others, incorporates topic-relatedness. Unlike these other studies however, \citet[p. 4756]{allaway-etal-2021-adversarial} ``do not assume a relationship between training and test topics'' as a fairer test of robustness. Results they present do show that stance detection is related to topic, but their efforts go to finding topic-invariant stance representations, which improves the generalizability of their model. 
Their consideration of topic similarity shows that topic difference is very relevant to stance detection.

\citet{aldayel_stance_2021} describe in their survey how several studies \cite{ klebanov2010vocabulary, zhu2019hierarchical, darwish2020unsupervised} show that texts pro or against an issue use different vocabularies (e.g.\ using `pro-life' when expressing a stance against abortion). Some of these studies attempt to leverage these vocabularies to generalize across similar topics. Recent work has looked into generalizing stance detection across datasets, task definitions, and domains \cite{schiller_stance_2021}, in which topic-specific performance is not mentioned. 

A recent approach to topic-specificity in stance detection is task adaptation. \citet{stein2021same} acknowledge that stance detection usually requires knowledge about the topic of discussion, which is not available for unseen topics. They approach this problem by changing the task to ``same-side stance classification", in which a model is trained to classify whether two arguments either have the same or a different stance. This reduces the model's leaning on topic-specific pro- and con-vocabulary, while still being able to separate different stances on the same topic. The best approach to this adapted task on a dedicated leaderboard\footnote{\url{https://webis.de/events/sameside-19/}, Accessed on the 22th of September 2021.} receives an F1 of .72 in the cross-topic setting with a fine-tuned BERT model \cite{ollinger_same_2020}.

Our current work adds the discussion of topic difference and topic specificity to state-of-the-art stance detection results. That is, earlier bag-of-words-based work considered lexical specificity of different topics for stance detection, and we add that into the discussion for the current state of the art: pre-trained, end-to-end neural models.

\section{Reproduction  Experiments}\label{sec:Repro}

\citet{reimers_classification_2019} apply their approach of cross-topic claim classification to two datasets: the \textit{UKP Sentential Argument Mining Corpus} \cite{stab2018argumentext} (`the UKP dataset') and the \textit{IBM Debater: Evidence Sentences} dataset \cite{shnarch-etal-2018-will} (`the IBM dataset'). We focus on the UKP Dataset, since the IBM Debater dataset has no `pro' and `con' class, but rather `evidence' and `no evidence' (and our focus is on stance detection).

As a second step after stance classification, the authors also attempt to cluster similar arguments within the same topic in a cross-topic training setting. We do not replicate this component, but instead dive deeper into the classification results.

We adopt the definition of reproduction by \citet{belz-etal-2021-systematic}: repeating the experiments as described in the earlier study, with the exact same data and software. We analyze our reproduced results according to the three dimensions of reproduction proposed by \citet{cohen2018three}: whether we find either the same or different (1) (numeric) values, (2) findings, and (3) conclusions as the earlier study.\footnote{For reasons of clarity, we present these dimensions in reverse order compared to \newcite{cohen2018three}. 
} Reproducing the same \textbf{values} means obtaining the same numeric results from a specific experiment. Experiments involving fine-tuning on BERT are non-deterministic. We therefore consider the metric fully reproduced if the original result lies within two standard deviations (stdevs) from our result, obtained from 10 random seeds.\footnote{The paper we reproduce, \citet{reimers_classification_2019}, does not provide model performance standard deviation over seeds.} The same \textbf{finding} means that the relation between the values associated with two or more dependent variables is the same, i.e.\ a system that outperformed another in the original study also does this in the reproduced study. The \textbf{conclusion} is the same when the broader implication of findings and values is the same. Conclusions are thus a matter of interpretation. As such, the same findings can lead to different conclusions and conclusions are, contrary to findings, not repeatable \cite{cohen2018three}. This section focuses on the repeatable components of reproducing a study: the values and the findings. We address the conclusions using our more detailed analyses in Section~\ref{sec:Topics}.

\subsection{Dataset Description}

The UKP dataset \cite{stab2018argumentext} consists of 25,492 argument sentences from 400 Internet texts (from essays to news texts) on 8 topics. 
The dataset designer's definition of claim is ``a span of text expressing evidence or reasoning that can be used to either support or oppose a given topic'' \cite[p. 3665] {stab2018argumentext}. They define topic as ``some matter of controversy for which there is an obvious polarity for possible outcomes'' \cite[p. 3665]{stab2018argumentext}, and map this polarity to a text expressing one of two classes: for or against the use, adoption, or idea of the topic under discussion. A third class is `no argument' to the topic under discussion, i.e.\ the text span falls outside of this polarity. 

The 8 topics in the dataset were randomly chosen from online lists of controversial topics on discussion websites \cite[p. 3666] {stab2018argumentext}. Specifically, these topics are \textit{abortion}, \textit{cloning}, \textit{death penalty}, \textit{gun control}, \textit{marijuana legalization}, \textit{minimum wage}, \textit{nuclear energy} and \textit{school uniforms}. The stance classes (pro, con, and no argument) were annotated by two argument mining experts and seven U.S.\ crowdworkers. The distribution of the dataset for different topics is shown in Table~\ref{tab:distribution}.
\begin{table*}[t]
\centering
\small{
\begin{tabular}{llccccccccc}
\hline
 & &  \small{abortion} & \small{cloning} & \small{death} & \small{gun} & \small{marijuana} & \small{minimum} & \small{nuclear} & \small{school} & \small{all}
\\
 &  & &  & \small{penalty} & \small{control} & \small{legalization} & \small{wage} & \small{energy} & \small{uniform} \\
\hline

\textbf{train} & \textit{pro} & 490 &  508 & 316 & 566 & 422 & 414 & 436 & 392 & 3.544\\
    & \textit{con} & 591 & 604 & 789 & 479 & 450  &  396 & 613 & 525 & 4.447\\
    & \textit{no arg} & 1.746 & 1.075 & 1.522 & 1.359 & 908 & 968 & 1.524 & 1.248 & 10.350\\
    \hline
\textbf{dev} & \textit{pro} & 54 & 56 & 38 & 63 & 47 & 46 & 48 & 44 & 396  \\
   & \textit{con} & 66 & 67 & 90 & 53 & 50 & 44 & 68 & 58 & 496 \\
   & \textit{no arg} & 195 & 120 & 165 & 152 & 101 & 108 & 170 & 139 & 1.150\\
   \hline
\textbf{test} & \textit{pro} & 136 & 142 & 103 & 158 & 118 & 116 & 122 & 109 & 1.004 \\
    & \textit{con} & 165 & 168 & 232 & 133 & 126 & 111 & 171 & 146 & 1.252 \\
    & \textit{no arg} & 486 & 299 & 396 & 378 & 253 & 270 & 424 & 347 & 2.853\\
\hline
\end{tabular}
}
\caption{Distribution of the UKP data over topics and over training (70\%), test (20\%), and validation (10\%) sets. 
}
\label{tab:distribution}
\end{table*}

In \citet{stab2018argumentext} we see a difference in agreement on stance classes in different topics, especially between expert and crowd. The topic achieving the highest agreement between crowd worker and expert is \textit{school uniforms} ($\kappa = .889$), and the lowest is \textit{death penalty} ($\kappa = .576$). The standard deviation over topics is .08 for expert--expert coded data and .16 for expert--crowd coded, both with a mean of $\kappa =.72$.

\subsection{Obtaining the Data}

The UKP Dataset is not available online due to copyright concerns, but there is a scraping script with archived hyperlinks available on \citet{reimers_classification_2019}'s GitHub page. 
We ran this script with all specifications given. The scraping script was able to return all claims on 6 of the 8 topics. The topics for which not all claims were detected were \textit{nuclear energy} and \textit{minimum wage}. We then instead obtained the complete datafiles from the authors.\footnote{These files revealed that the scraping script broke down in the \textit{minimum wage} topic due to one specific claim that was archived, but could not be retrieved. 
``Despite the inevitable negative outcomes that will surely result from a \$ 15 minimum wage – we ’ve already seen negative effects in Seattle ’s restaurant industry – politicians and unions seem intent on engaging in an activity that could be described as an “economic death wish.'' We speculate this claim could possibly not be retrieved due to it containing the dollar sign, \url{https://web.archive.org/web/20160217041546/http://www.aei.org:80/publication/ten-reasons-economists-object-to-the-minimum-wage/}}

\subsection{Training and Evaluation Method}

\citet{reimers_classification_2019} use the training method described in \citet{stab2018argumentext}. Each topic is split into a training (70\%), development (10\%), and test split (20\%). Training is done on the training splits of 7 topics, tuned on the development split (10\%) of these 7 topics, and finally evaluated on the test split (20\%) of the held-out 8th topic. They do this for each of the 8 topics (holding out a different topic each time), then apply this procedure for 10 different random seeds on a GPU. Evaluation is assessed with macro F1, averaged over all topics and all random seeds. Their best performing model is a fine-tuned BERT-large model \cite{devlin-etal-2019-bert}, but with only minor improvement over BERT-base.

We use the same training set-up and BERT models for our reproduction. For training, we use the author's code with Python3.8 on a single NVIDIA GeForce RTX 2080 Ti GPU. Our learning rate is 2e-5 for both models, as in \citet{reimers_classification_2019}.\footnote{All our code can be found in the following GitHub repository: \url{https://github.com/myrthereuver/claims-reproduction}.}

We additionally train a non-BERT model (a Support Vector Machine (SVM) with tf-idf features) in the same hold-one-topic-out manner. Tf-idf-based approaches have shown quite solid performance on stance detection in prior work \cite{riedel2017simple}. This model is deterministic and is thus not run with multiple seeds. It is run with Python3.9 and the sklearn package. The SVM is intended for the feature analysis in Section~\ref{sec:Analysis}, but we present the performance of this model also in Table~\ref{tab:reproduction} and the following section.

\subsection{Results of Reproduction}

\begin{table*}[h]
\centering
\small{
\begin{tabular}{l|cccccccc}
\hline
\textbf{Model}  & \multicolumn{6}{c}{\textbf{UKP Dataset}} \\
\hline
 \small{mean (stdev) 10 seeds} & F1 & P \small{pro} & P \small{con} & R \small{pro} & R \small{con} 
\\\hline
   \small{\citet{reimers_classification_2019} biclstm+BERT} & .424 & .267 & .389 & .281 & .403  \\
  \small{\citet{reimers_classification_2019} BERT base} & .613 (-) & .505 (-) & .531 (-) & .470 (-) & .576 (-)  \\
   \small{\citet{reimers_classification_2019} BERT large} & \textbf{.633} (-) & .554 (-) & .584 (-) & .505 (-) & .560 (-) \\
 \hline
 \hline
   \small{SVM+tf-idf} & .517 & .418 & .460 & .414 &  .423 & \\
 \small{Reproduction BERT-base} & \textbf{\small{.617 (.006)}} & .519 (.011) & .538 (.007) & .464 (.029) & .581 (.019) &  \\
  \small{Repr. BERT-large - all seeds} & .596 (.043) & .483 (.057) & .527 (.057) & .464 (.058) & .516 (.063) \\ \hline \hline
\small{Repr. BERT-large - 5 evenly performing seeds} & .636 (.007) & .532 (.014) & .578 (.016) & .515 (.016) & .567 (.022)  \\ \hline
 
\end{tabular}}
\caption{Reproduction results \citet{reimers_classification_2019}. The fourth row shows our non-BERT model (an SVM) beating their LSTM baseline, and the fourth and fifth row show the results of our BERT reproductions. The sixth row shows an average BERT-large performance without the 5 seeds that considerably under-performed for one topic.
}
\label{tab:reproduction}
\end{table*}

\paragraph{BERT-base} Table \ref{tab:reproduction} shows that mean performance over the 3 classes (`pro', `con', or `no argument') is F1 = .617 (stdev over 10 seeds = .006). \citet{reimers_classification_2019}'s reported result (F1 = .613) lies within 1 stdev from this result.

\paragraph{BERT-large} Mean performance over all topics and stance classes is F1 = .596 (stdev over 10 seeds = .043). The performance reported in \citet{reimers_classification_2019} is F1 = .633, which lies within 2 stdev of our result. However, our stdev is relatively high due to high variance of performance over different seeds, with half of our seeds performing noticeably lower than even BERT-base.\footnote{Our large variance in performance over seeds is due to each seed fine-tuning the model 8 times (once for each topic). The 5 unevenly performing seeds each under-perform on a different topic (F1 < .50) due to only assigning the majority class (`no argument'). Other topics in these 5 seeds do outperform BERT-base.} For the other 5 seeds, the model performed better (F1 = .636, stdev = .007), and within one (much smaller) stdev of the performance reported in \citet{reimers_classification_2019}.

\paragraph{SVM+tf-idf (non-BERT model)} This model performs at F1 = .517 averaged over the held-out topics and three classes (`pro', `con', and `no argument'), see Table \ref{tab:reproduction}. This outperforms by .10 points in F1 the best performing LSTM-based architecture presented in \citet{stab2018argumentext} (F1 = .424), a baseline in \citet{reimers_classification_2019}. Their performance improvement of the BERT model over LSTM was .20 in F1. Comparing our SVM model to BERT, we find a smaller improvement over a non-BERT model: .10 F1 improvement for BERT-base (F1 = .617). 
Our BERT models still outperform our non-BERT model, as in \citet{reimers_classification_2019}. Our SVM result does fall within 2 stdevs of BERT-large, but this is due to BERT-large's substantial stdev due to a steep drop in performance for half of the seeds.

\subsection{Conclusion of reproduction}

\citet{reimers_classification_2019}'s results are reproducible in the sense the first dimension of reproducibility \cite{cohen2018three}: the originally reported numeric \textbf{values} fell within 2 stdevs of our reproduced results for both BERT-base and BERT-large. For BERT-base and 5 of the 10 seeds in BERT-large, we obtained a precision, recall, and F1 that are very similar to the original study.

The results are also reproducible in four of the five reproducibility aspects identified by \citet{fokkens2013offspring}: under-descriptions of preprocessing, experimental set-up, versioning, and system output. These were described in either the paper, on the author's GitHub page, or in code documentation. We do observe differences in relation to  `system variation' which is inherent to training neural networks, where identical results are seldom obtained. These variations were small for most experiments, except for the 5 random seeds that led to substantial under-performing on one topic for BERT-large.

When looking at the second dimension of reproducibility defined by 
\citet{cohen2018three} (\textbf{findings}), we observe that BERT-base and BERT-large indeed clearly outperform the LSTM baselines from \citet{stab2018argumentext} as well as our own stronger SVM+tf-idf non-BERT model on the stance detection task. We were able to reproduce the reported increase in performance of BERT-large over BERT-base and non-BERT models. However, BERT-large also showed considerable under-performance on one topic in 5 out of 10 seeds. We see this outcome as a confirmation that it is important to look at different seeds, and that care should be taken when drawing conclusions based on minor differences when working with neural models.

The third dimension of reproducibility is that of \textbf{conclusions}. \newcite{reimers_classification_2019} conclude that BERT strongly outperforms previous results on identifying arguments for unseen topics, which we confirm, and that these results are ``very encouraging and stress the feasibility of the task'' \cite[p. 575]{reimers_classification_2019}. The remainder of this paper provides further analyses to investigate whether our results also lead to this overall conclusion. In particular, we investigate how our models perform on individual topics (Section~\ref{sec:Topics}) and generic topic-independent signals in the data (Section~\ref{sec:Analysis}).
\newpage
\section{Topic Specifics in Classification}\label{sec:Topics}

To support the conclusions in \citet{reimers_classification_2019} on the success of cross-topic stance detection, we expect a relative stability of performance over topics. The following sections go into some details not explored in \citet{reimers_classification_2019}, specifically the cross-topic performance of different topics, and the interaction between topic and class and its influence on performance.

\subsection{Variance over (classes in) topics}

\begin{table*}[h]
\centering
\small{
\begin{tabular}{l|cccccccc}
\hline
\small{\textbf{held-out}} & \small{abortion} & \small{cloning} & \small{death} & \small{gun} & \small{marijuana} & \small{minimum} & \small{nuclear} & \small{school} 
\\
\small{\textbf{topic}} &  &  & \small{penalty} & \small{control} & \small{legalization} & \small{wage} & \small{energy} & \small{uniform} \\
\hline
\small{SVM+tf-idf} & \small{.463} & \small{.585}  & \small{.482} & \small{.515} & \small{.323} & \small{.615} & \small{.598} & \small{.576}   \\
 
\small{BERT-base} & \small{.533 (.011)} & \small{.693 (.013)} & \small{.562 (.012)} & \small{.530 (.013)} & \small{.607 (.016)} & \small{.670 (.009)} & \small{.660 (.011)} & \small{.678 (.016)} \\

\small{diff. } &
\small{+.070} & \small{+.108}  & \small{+.080} & \small{\textit{+.028}} & \small{\textbf{+.283}} & \small{+.055} & \small{+.0850} & \small{+.102}   \\
\hline
\end{tabular}
}
\caption{BERT-base's performance in F1 (macro) on different held-out topics. The \textit{italicized} difference shows the smallest difference between the SVM model and the BERT-base model (on the gun control topic), while the \textbf{bolded} difference shows the largest difference (on the marijuana topic).}
\label{tab:reproduction_topics}
\end{table*}

Table~\ref{tab:reproduction_topics} presents the performance of the models on individual topics. The results show that some topics perform considerably worse than others with the cross-topic training method (training on seven topics and testing on the held-out eighth topic). The \textit{cloning} topic performs more than .07 F1 higher than the averaged model performance (F1 = .693 vs F1 = .617). The \textit{abortion} and \textit{gun control} topics perform almost .09 lower than the averaged model performance (F1 = .533 \& .530 vs F1 = .617). Note that a difference nearing .10 in F1 score is relatively large, as it is comparable to the difference between the SVM performance and the state-of-the-art BERT models in the previous section.

A per-topic analysis in Table \ref{tab:reproduction_topics} shows that the SVM+tf-idf model performs within .10 points of the BERT-base model for seven of the eight topics, with some performing less than .3 points lower than BERT. The only exception is the topic \textit{marijuana legalization}, which performs .28 points lower than the BERT model. The large average performance increase (+.11 in F1) over SVM comes from BERT-base improving performance on this one topic.

\begin{figure*}[h!]
\centering
\includegraphics[width=0.90\textwidth,trim={0 2cm 0 0},clip]{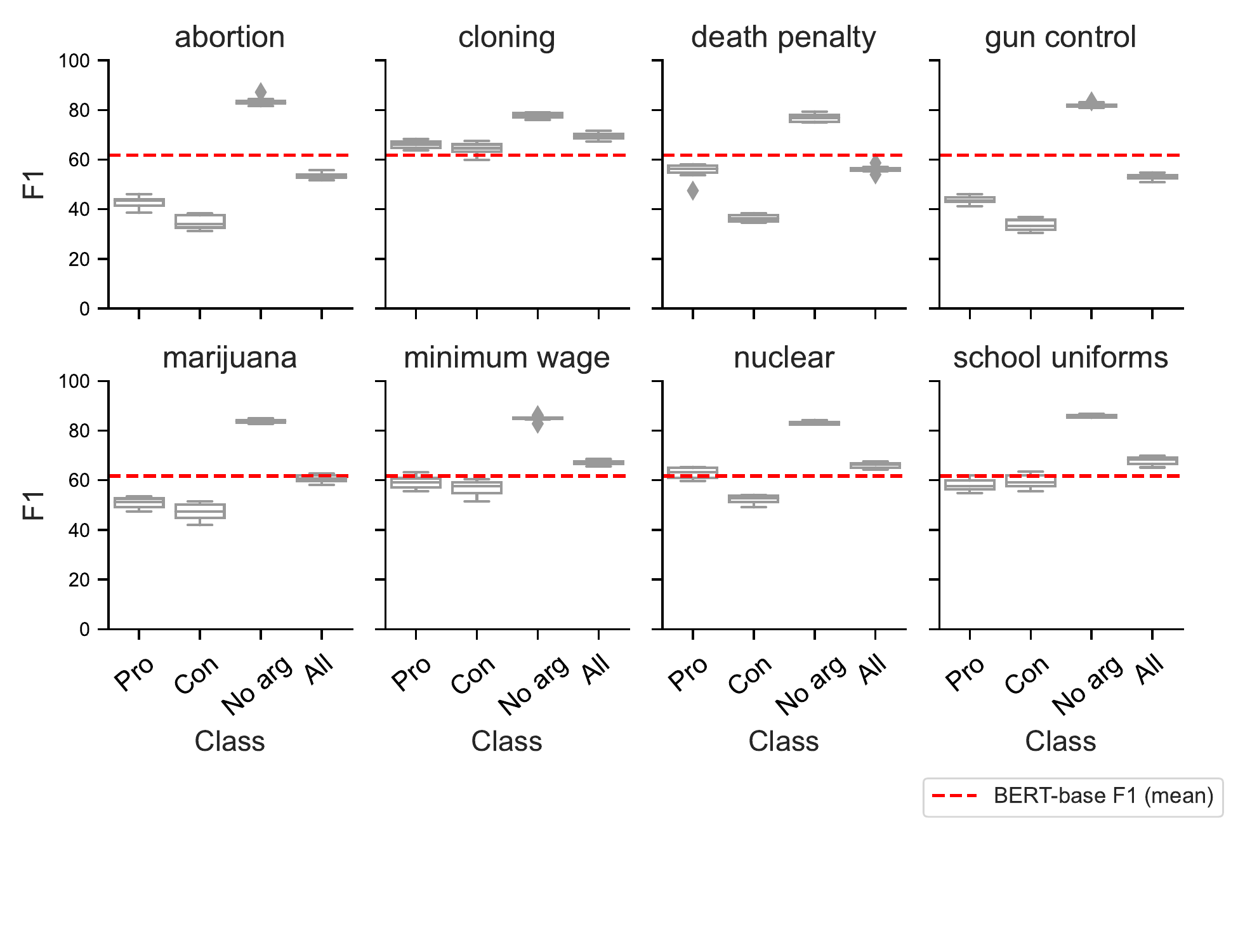}
\caption{BERT-base's performance on different topics plotted in a boxplot, with on the y-axis the F1 score of the 4 categories plotted on the x-axis: `pro', `con', `no argument', and overall. A longer boxplots means more variability over seeds in score. The red line represent the averaged F1 score of the same model (BERT-base), presented as model performance in \citet{reimers_classification_2019}.}
\label{fig:topic/class}
\end{figure*}

Figure~\ref{fig:topic/class} presents the BERT-base in-class F1 score of the three classes (`pro', `con', `no argument'), and in-topic averaged F1. The red line indicates the average model performance of .617. We see some consistency, e.g.\ the `no argument' class consistently scoring around F1 = .80, but we also see some topic-specific behavior. \textit{Cloning}, \textit{minimum wage}, and \textit{school uniforms} obtain higher F1 performance than average for all classes. In contrast, \textit{death penalty}, \textit{gun control}, and \textit{abortion} perform considerably lower than the average F1 performance in the `pro' and `con' classes. These topics see in-class performance of even F1 < .50.

Each cross-topic model is trained by removing one topic from the training data. In this way, we remove a different number of training examples each time. The topics with the most training examples for a class (e.g.\ `pro' in the \textit{gun control} topic) therefore have a smaller training set for this class when training a cross-topic model. If there were a linear relationship between dataset size and performance, one would expect that topics with \textit{fewer} training examples (and therefore more training examples left when this topic is left out of training) to do better than topics with \textit{more} training examples (whose cross-topic models lose more training examples). 
        
Table~\ref{tab:distribution} does show that the `no argument' class has a three times larger proportion of the training set than the `pro' and `con' classes, which could explain the better performance of this class in all topics, but training set size difference does not account for the between-topic variation in the `pro' and `con' classes. Instead, Table~\ref{tab:distribution} shows that topics with the most training examples (that means, the largest set of examples removed in a cross-topic model) do not have the worst performing cross-topic models in Figure~\ref{fig:topic/class}. For example, the \textit{abortion} topic has relatively few `con' examples removed (591) compared to other classes such as \textit{cloning}, \textit{death penalty}, and \textit{nuclear energy}, and yet has the lowest in-class F1 for the `con' class (in-class F1 = .40). Performance thus appears to be less related to the number of training examples.

\begin{table*}[t]
\centering
\begin{tabular}{lllll|lllll}
\hline
\small{all topics} & & & & & \small{abortion topic}\\ \hline
 \small{Pro (vs Con)} &\small{Con (vs Pro)} & \multicolumn{2}{c}{\small{No Argument}} & & \small{Pro (vs Con)} &\small{Con (vs Pro)} & \multicolumn{2}{c}{\small{No Argument}} \\\hline
  \small{}  & &  \small{vs Pro} & \small{vs Con}  &  & & & \small{vs Pro} & \small{vs Con} \\\hline 
 \small{pejorative} & \small{\textit{morality}} & \small{basic} & \small{pronounced} &  & \small{seek} & \small{babies} & \small{way} & \small{anti}  \\
 
 \small{pronounced} & \small{format} & \small{section} & \small{threatens} & &  \small{illegal} & \small{abortion} & \small{against} & \small{ways}  \\ 
 
 \small{activity} & \small{\textit{bill}} & \small{take} & \small{additional} & & \small{reproductive} & \small{life} & \small{we} & \small{over}  \\

  \small{relations} & \small{workshop} & \small{\textbf{robert}} & \small{revolt} &  & \small{simply} & \small{conception} & \small{side} & \small{always}  \\

   \small{additional} & \small{workers} & \small{introduced} & \small{now}   & & \small{humane} & \small{simply} & \small{justify} & \small{thing}  \\
   
    \small{unexceptional} & \small{\textit{sources}} & \small{unquestioned} & \small{proper}  &   & \small{bear} & \small{risks} & \small{experience} & \small{question} \\
    
  \small{threatens} & \small{\textit{philosophical}} & \small{revolt} & \small{typical}  &  & \small{lifers} & \small{abortions} & \small{held} & \small{performed}  \\

 \small{variable} & \small{coincidentally} & \small{scientifically} & \small{mentor} & & \small{mother} & \small{complications} & \small{tell} & \small{debate}  \\
 
 \small{39th} & \small{\textit{statutes}} & \small{\textbf{lifenews}} & \small{sharing} & & \small{healthy} & \small{birth} & \small{single} & \small{illegal}  \\
  \small{where} & \small{phrases} & \small{individuals} & \small{denuded} &  & \small{lives} & \small{kill} & \small{had} & \small{equal}   \\
  \hline
\end{tabular}
\caption{Top-features for different topics according to SVM, Pairwise F-based feature analysis. We see potentially meaningful words in \textit{italics} (the `con' class has features based on morality and legality, e.g.\ bills and statutes), and potential spurious features in \textbf{bold} (such as names websites and even of individuals).\\}
\label{tab:repr_results_SVM_features}
\end{table*}

We investigated the source of low performance on the `pro' and `con' class in the \textit{abortion} topic with confusion matrices, and compared this to a topic where pro and against did not under-perform (\textit{minimum wage}). We did not pick one specific seed, but calculated the mean percentage of `true' examples in each confusion matrix cell over all 10 seeds. 
In the \textit{abortion} topic, 44 \% of `pro' arguments get classified as `against', and only 33\% get correctly classified as `pro'. The \textit{minimum wage} topic shows no discernible pro/against classification confusion, and 60\% of all true `pro' and `against' arguments are correctly classified. The section below analyzes the misclassifications in low-performing topics. 

\subsection{Qualitative Analysis of Misclassification}

The low performance of `pro' and `con' in some topics (\textit{abortion}, \textit{gun control}, and \textit{death penalty}) warrants some further investigation. Table \ref{tab:repr_missclass} shows 
four example misclassifications between `pro' and `con' by BERT-large in the test examples the model encountered on these topics.\footnote{To ensure we are not cherry-picking examples, we looked at errors that were not unique to just one seed, and identified these examples as salient examples of a general trend.}

\begin{table*}[h!]
\centering
\small{
\begin{tabularx}{\linewidth}{lllll}
\hline
\small{\textbf{Topic}} & \small{\textbf{True}} & \textbf{Pred} &\textbf{Sentence} & \textbf{Frequency in seeds}\\
\hline
\small{gun contr.} & pro & con & "When high-capacity magazines were used in mass shootings, & 9/10 \\ &&& the death rate rose 63 \% \\ &&& and the injury rate  rose 156 \% ." \\
\small{gun contr.} & con & pro & "[..] The Second Amendment   protects an individual right & 7/10 \\ &&&  to possess a firearm  unconnected \\ &&& with service in a militia , and to use that arm for traditionally lawful purposes  \\ &&& , such as self-defense within the home . "\\
\small{gun contr.} & pro & con & "In this crossfire , bullets would likely hit civilians & 9/10 \\ &&& ( imagine a room filled with a crowd and three people shooting \\ &&& at each other ) and the casualty count would increase." \\
\small{gun contr.} & con & pro & "Gun enthusiasts understand the benefit & 7/10 \\ &&& of large ammo feeders and wish to defend them because  \\ &&& they recognize the advantage that such feeders give."\\ 
\hline
abortion & pro & con & "Not only has the biological development not yet occurred to  & 4/10 \\ &&& support  pain experience ,   but the environment after birth , so necessary \\\ &&& to the development of pain experience , is also yet to occur ." \\
abortion & pro & con & "Warren concludes that as the fetus satisfies only one criterion, & 5/10 \\&&& consciousness ( and this only after it becomes susceptible to pain )\\&&& , the fetus is not a person  and abortion is therefore morally permissible ."\\
abortion & con & pro & It is argued that just as it would not be permissible to refuse & 2/10 \\ &&& temporary accommodation \\ &&& for the guest to protect him from physical harm , \\ &&& it would not be permissible to refuse  temporary accommodation of a fetus .\\
abortion & con & pro & "92 \% of abortions in America are purely elective & 3/10 \\ &&& -- done on healthy women to end the lives \\ &&& of healthy children." \\

\hline

death pen. & con & pro & Mentally ill patients may be put to death .& 2/10 \\
death pen. & con & pro & Evidence shows execution does not act as a deterrent to capital punishment. & 9/10\\
death pen. & pro & con & A system in place for the purpose & 8/10 \\&&& of granting justice can not do so for the surviving victims , \\ &&& unless the murderer himself is put to death . \\
death pen. & con & pro & CON : " ... Since the reinstatement of the modern death pen. , & 9 /10 \\&&& 87 people have been freed from death row because \\&&& they were later proven innocent . \\
\hline
\end{tabularx}
}
\caption{Misclassifications on political topics with considerable `pro' and `con' confusion}
\label{tab:repr_missclass}
\end{table*}

We find two types of misclassifications, each related to topic-specific differences to stance classes. The first type is {\bf misclassification due to the socio-cultural background knowledge and context} of a specific topic's 
arguments. The second type is related to a model taking the {\bf stance towards a subcomponent of a topic} and confusing it for the text's overall stance on the topic, e.g.\ statements in the `pro' class mostly expressing views \textit{against} something else related to the argument (unwanted pregnancies, gun violence, innocents dying). 

Examples of both issues are arguments centering around ``many innocents (babies, children, mentally ill) will die". 
There are 5 variations of this argument in these 3 topics: row 1 and row 3 (\textit{gun control}), row 8 (\textit{abortion}), and rows 9 and 12 (\textit{death penalty}) in Table~\ref{tab:repr_missclass}. Not only is one usage of this argument traditionally connected to the `pro' class of one topic (\textit{gun control}), and the `con' class of another (\textit{abortion}), the implication is: innocents dying is bad. The model seems to lack this world knowledge, and for instance classifies this argument as `pro' \textit{death penalty}.

Another salient example is row 2 of Table~\ref{tab:repr_missclass}. This argument argues \textit{in favor} of gun rights for self-defense, but the model misclassifies this as \textit{against} gun control. The model also fails to connect the second amendment discussion to the \textit{against gun control} class. This is the same mistake made by the LSTM-model in \citet[p.3671]{stab2018argumentext}, showing that BERT appears to not improve over LSTM on the topic-specific nuances here. In other words, it fails to correctly identify the socio-cultural dimensions \cite{du2007stance} of this debate.

\newpage
~\newpage
\newpage
~\newpage

\subsection{SVM and Lexical Features}\label{sec:Analysis}

To analyze which words are used in relation to specific stances and topics, we trained an SVM model with tf-idf features on stance detection on all topics (F1 = .573). For each class pair (`pro' vs `con', `pro' vs `no-argument', etc.), we extracted top-10 features with the highest coefficient for that specific class. 

Table~\ref{tab:repr_results_SVM_features} presents the most important features of the topic-agnostic model trained on all topics. Some unigrams 
appear meaningful for the class. For instance, in the cross-topic setting, the word ``morality'' is a feature for the `con' class. In contrast, the `no argument' class is often identified with words that appear to have little content-relationship to the class identity: a topic-specific pro-life website (lifenews) or someone's name (`robert'). 

We also trained within-topic models to find whether there is topic-specific vocabulary related to stance that differs from the topic-agnostic model. Table \ref{tab:repr_results_SVM_features} also presents the 10 most informative features for a model trained on only the abortion topic (F1 = .595). Immediately we see that there is only limited overlap with the lexical features used to decide between `pro' and `con' in a multi-topic scenario. Within only the abortion topic, the `pro' and `con' class are defined by concepts related to the lexical content of this specific discussion: babies, life, and birth. We also see the contrast between `pro' arguments talking about reproduction and the mother, while the `con' arguments mention life, conception, and babies. This lexical feature analysis shows no apparent overlap between the topic-specific features in the abortion model and the topic-independent features in the topic-agnostic model. This might indicate that vocabulary is quite specifically related to topics in stance detection.

\section{Conclusion: Topic Matters} \label{sec:Discussion}
Stance detection is a difficult NLP task. Despite recent advances by pre-trained Transformers, these models have similar issues in a cross-topic setting as earlier models. This paper reproduced 
stance detection experiments with pre-trained Transformers by \citet{reimers_classification_2019}, training on seven topics and testing on an eighth topic. We found similar results, but also both class and topic influencing performance. Cross-topic BERT models perform below mean model performance in some topics (\textit{abortion}, \textit{gun control}) on the pro and con classes. 

This makes us pause about \citet{reimers_classification_2019}'s main claim: does BERT improve cross-topic stance detection over non-Transformer models? We argue this claim needs an asterisk: this cross-topic approach does not work as well for all topics. Different topics show specific vocabularies and socio-cultural contexts, and especially these specific contexts BERT cannot navigate. BERT models still make similar mistakes on gun control as the LSTM-based models in \citet{stab2018argumentext}. 

These findings lead us to two take-aways. Firstly, we hypothesize that models like BERT rely more on topic-specific features for stance detection than topic-independent lexical words related to argumentation. \citet{thorn-jakobsen-etal-2021-spurious} also recently found this, and connected BERT's cross-topic stance detection performance to its focus on spurious topic-specific lexical features ("gun", "criminal") rather than words related to argumentation. They also conclude a fair real-world evaluation of cross-topic stance detection means reporting the worst performing cross-topic pair rather than average performance over topics. 

Secondly, we also think it is necessary to analyze the context of topics, and its relation to other debate topics within and outside the dataset. Most topics in stance detection studies are currently U.S. socio-political issues. This goes beyond a limitation of language, such as a focus on English without specifying this \cite{bender2019rule}, since the same socio-cultural topics are not even universally relevant in the English-speaking world (gun control is not a salient discussion in Scotland). Such a focus on topic diversity is also important for use-cases. For diversity of viewpoints in search \cite{draws2021assessing} or news recommendation \cite{reuver-etal-2021-nlp}, stance detection needs to work on many different topics.

\citet{schlangen-20191-natural} states that we need to carefully define specific NLP tasks and capabilities needed to solve them. Modelling cross-topic stance detection in a topic-agnostic manner, while divorcing it from socio-cultural context, might not do justice to stance detection. Future work might focus on the specifics of topics: analyzing similarity between discussions \cite{wei2019modeling}, or modelling required socio-cultural contextual knowledge (`second amendment is related to gun control'). Models able to deal with topic-specific vocabulary and socio-cultural context of debates might improve on the state-of-the-art of cross-topic stance detection.

\newpage
\section*{Acknowledgments}
This research is funded through Open Competition Digitalization Humanities and Social Science grant nr 406.D1.19.073 awarded by the Netherlands Organization of Scientific Research (NWO). Our computing was done through SURF Research Cloud, a national supercomputer infrastructure in the Netherlands also funded by the NWO. We would like to thank dr. Nils Reimers for sending us their paper's data. We would also like to thank the anonymous reviewers, whose very helpful comments improved the paper. All opinions and remaining errors are our own. 

\bibliography{anthology,ClaimsMay2021.bib}
\bibliographystyle{acl_natbib}

\appendix

\newpage

\end{document}